\documentclass[10pt,twocolumn,letterpaper]{article}

\usepackage{cvpr}
\usepackage{times}
\usepackage{epsfig}
\usepackage{graphicx}
\usepackage{amsmath}
\usepackage{amssymb}
\usepackage{makecell}
\usepackage{float}
\usepackage{subcaption}
\usepackage{multirow}
\usepackage{bm}
\usepackage{float}
\usepackage{textcomp}
\usepackage{gensymb}

\usepackage{enumitem}
\usepackage{mathrsfs}
\usepackage[dvipsnames]{xcolor}

\newcolumntype{L}[1]{>{\raggedright\arraybackslash}p{#1}}
\newcolumntype{C}[1]{>{\centering\arraybackslash}p{#1}}
\newcolumntype{R}[1]{>{\raggedleft\arraybackslash}p{#1}}

\restylefloat{table}

\usepackage[pagebackref=true,breaklinks=true,letterpaper=true,colorlinks,bookmarks=false]{hyperref}

\cvprfinalcopy 

\def\cvprPaperID{325} 

\ifcvprfinal\fi
\begin{document}

\title{ContextDesc: Local Descriptor Augmentation with Cross-Modality Context}

\author{Zixin Luo$^{1}$\hspace{0.7cm} Tianwei Shen$^{1}$\hspace{0.7cm} Lei Zhou$^{1}$\hspace{0.7cm} Jiahui Zhang$^{2}$ \\ Yao Yao$^{1}$\hspace{0.7cm} Shiwei Li$^{1}$\hspace{0.7cm} Tian Fang$^{3}$\hspace{0.7cm} Long Quan$^{1}$ \\
\normalsize $^1$Hong Kong University of Science and Technology \\ \normalsize $^2$Tsinghua University \hspace{0.7cm} \normalsize $^3$Shenzhen Zhuke Innovation Technology (Altizure) \\
\tt\small\{zluoag,tshenaa,lzhouai,yyaoag,slibc,quan\}@cse.ust.hk \\
\tt\small jiahui-z15@mails.tsinghua.edu.cn\hspace{0.7cm} fangtian@altizure.com}

\maketitle
\thispagestyle{empty}

\begin{abstract}
Most existing studies on learning local features focus on the patch-based descriptions of individual keypoints, whereas neglecting the spatial relations established from their keypoint locations. In this paper, we go beyond the local detail representation by introducing context awareness to augment off-the-shelf local feature descriptors. Specifically, we propose a unified learning framework that leverages and aggregates the cross-modality contextual information, including (i) visual context from high-level image representation, and (ii) geometric context from 2D keypoint distribution. Moreover, we propose an effective N-pair loss that eschews the empirical hyper-parameter search and improves the convergence. The proposed augmentation scheme is lightweight compared with the raw local feature description, meanwhile improves remarkably on several large-scale benchmarks with diversified scenes, which demonstrates both strong practicality and generalization ability in geometric matching applications. [\href{https://github.com/lzx551402/contextdesc}{code release}]
\end{abstract}
\section{Introduction}

\begin{figure}[t]
	\centering 
	\begin{subfigure}[t]{0.48\textwidth}
		\includegraphics[width=\textwidth]{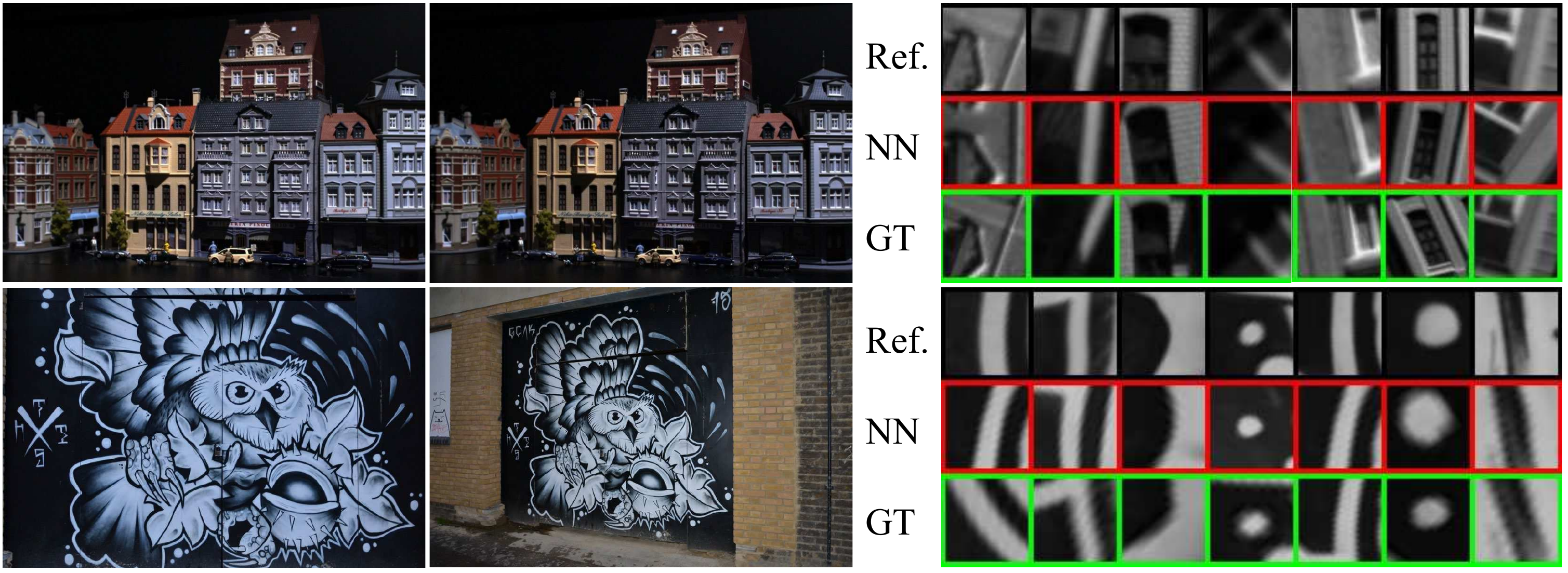}
		\vspace{-5mm}
		\caption{}
		\label{fig:saturate}
	\end{subfigure}
	\begin{subfigure}[t]{0.48\textwidth}
		\includegraphics[width=\textwidth]{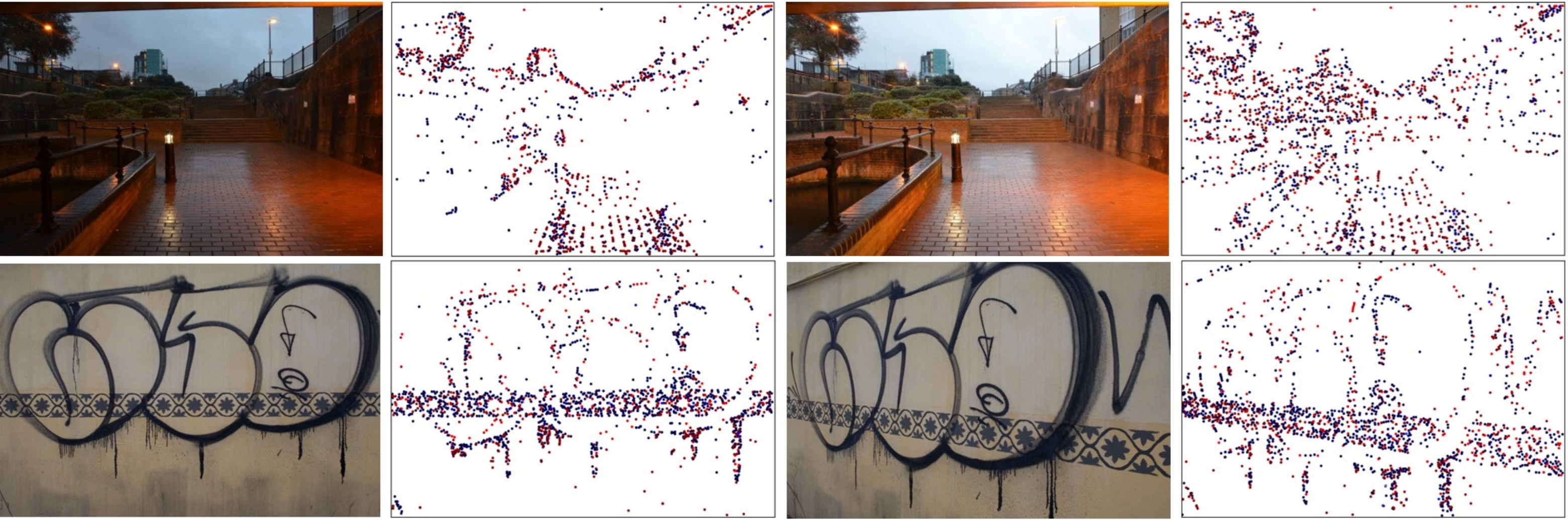}
		\vspace{-5mm}
		\caption{}
		\label{fig:geo_structure_example}
	\end{subfigure}
	\vspace{-4mm}
	\caption{(a) Saturated results on standard benchmark~\cite{balntas2017hpatches} by a recent method~\cite{Luo_2018_ECCV}. The search of nearest neighbors (NN) returns false matches though visually similar to groundtruth (GT), indicating the limitation of relying on only local visual information. (b) 2D keypoints distribute structurally, on which we human beings are capable of establishing coarse matches even without color information.}
	\vspace{-4mm}
\end{figure}

Designing powerful local feature descriptor is a fundamental problem in applications such as panorama stitching~\cite{li2015dual}, wide-baseline matching~\cite{matas2004robust,zhou2018learning,zhou2017progressive}, image retrieval~\cite{nister2006scalable} and structure-from-motion (SfM)~\cite{Zhu_2018_CVPR,shen2016graph,zhang2017distributed,zhu2014local}.
Despite the recent notable achievements, the performance of state-of-the-art learned descriptors is observed to be somewhat saturated on standard benchmarks. As shown in Fig.~\ref{fig:saturate}, due to repetitive patterns, the matching algorithm often finds false matches as nearest neighbors that are visually indistinguishable from groundtruth, unless validated by geometry. Essentially, such visual ambiguity may not be easily resolved given only local information. In this spirit, we seek to enhance the local feature description with extra prior knowledge, which we refer to as introducing \emph{context awareness} to augment local feature descriptors. 

As a common practice, a multi-scale-like architecture can help to capture \emph{visual context} of different levels, which is referred to as multi-scale domain aggregation by DSP-SIFT~\cite{dong2015domain} and adopted by recent learned descriptors~\cite{zagoruyko2015learning,kumar2016learning,tian2017l2}. Beside of the challenge on selecting proper domain sizes, a na\"ive multi-scale implementation may cost excessive computation such as doubled inference time and doubled feature dimensionality~\cite{zagoruyko2015learning,kumar2016learning,tian2017l2}. Seeking for more reasonable accuracy-efficiency trade-offs, we instead resort to well-studied high-level image representation, e.g., the regional representation used by image retrieval studies~\cite{radenovic2016cnn,shen2018matchable} which essentially incorporates rich image context. Thereby, we strive to effectively combine the local feature description and off-the-shelf visual understandings so as to go beyond the local detail representation.


In addition, it would be interesting to exploit context in other modality. In particular, as shown in Fig.~\ref{fig:geo_structure_example}, since keypoint is principally designed to be repeatable in the same underlying scene, its distribution thus reveals comprehensive scene structure that allows we human beings to establish coarse matches even without color information, which further enables us to explore \emph{geometric context} formed by the spatial relations of keypoints to help to alleviate the visual ambiguity of local descriptions. 

Thus far, we have discussed two context candidates, referred to as \emph{visual context} and \emph{geometric context} that incorporate high-level visual representation over the image and geometric cues from 2D keypoint distribution, respectively. Instead of learning a completely new descriptor, in the present work, we target to flexibly leverage the above context awareness to augment off-the-shelf local descriptors without altering their dimensionality, in which process we consider the key challenges threefold: 

\vspace{-1mm}
\begin{itemize}[leftmargin=*]\itemsep0em
	\item A proper integration of geometric local feature and semantic high-level representation. As keypoint description requires sub-pixel accuracy, the integration is not supposed to obscure the raw representation of local details.
	
	\item The instability of 2D keypoint distribution. Due to image appearance changes, keypoint distribution often suffers from substantial variations of sparsity, non-uniformity or perspective, which raises difficulties on acquiring strong invariance property of the feature encoder.
	
	\item An effective learning scheme. Input signals and features in different modalities are supposed to be efficiently processed and aggregated in a unified framework. 
\end{itemize}
\vspace{-1mm}

Finally, regarding practicability, the augmentation is not supposed to introduce excessive computational cost, as the local feature description is often regarded as part of preprocessing in practical pipelines. 

Although contextual information has been widely explored in semantic-based tasks, the challenges faced by local feature learning are substantially different, posing many non-trivial technical and systematic issues to overcome. In this paper, we propose a unified augmentation scheme that effectively leverages and aggregates cross-modality context, of which the contributions are summarized threefold: 1) a novel \emph{visual context encoder} that integrates high-level visual understandings from \emph{regional image representation}, a technique often used by image retrieval~\cite{radenovic2016cnn,shen2018matchable}. 2) A novel \emph{geometric context encoder} that consumes unordered points and exploits geometric cues from 2D keypoint distribution, while being robust to complex variations. 3) A novel N-pair loss that requires no manual hyper-parameter search and has better convergence properties. To our best knowledge, it is the first work that emphasizes the importance of context awareness, and in particular addresses the usability of spatial relations of keypoints in local feature learning.

The proposed augmentation is extensively evaluated and achieves state-of-the-art results on several large-scale benchmarks, including patch-level homography dataset, image-level wild outdoor/indoor scenes and application-level 3D reconstruction image sets, while being lightweight compared with raw local description, demonstrating both strong generalization ability and practicability.


\section{Related Work}
\label{sec:related_work}
\paragraph{Learned local descriptors.}
Initially, local descriptors are jointly learned with a new comparison metric~\cite{han2015matchnet,zagoruyko2015learning}, which is later simplified as direct comparison in Euclidean space~\cite{simo2015discriminative,yi2016lift,balntas2016learning,kumar2016learning,balntas2016pn}. More recently, efforts are spent on efficient training data sampling~\cite{tian2017l2,mishchuk2017working,he2018local}, effective regularizations~\cite{tian2017l2,zhang2017learning}, and geometric shape estimation of input patches~\cite{moo2016learning,AffNet2017}. However, most of above methods take individual image patches as input, whereas in the present work, we aim to take advantage of contextual cues beyond the local detail and incorporate features in multiple modalities.

\smallskip\noindent\textbf{Context awareness.}
Although widely introduced in computer vision tasks, context awareness has received little attention in learning 2D local descriptors. In terms of visual context, the central-surround (CS) structure~\cite{zagoruyko2015learning,kumar2016learning,tian2017l2} leverages multi-scale information by additionally feeding the central part of patches to boost the performance, whereas sacrificing computational efficiency due to the doubled extraction time and feature dimensionality. To incorporate semantics, one previous practice~\cite{kobyshev2014matching} designs a new comparison metric and describes features from histogram of semantic labels. In contrast to geometric matching, a family of studies has focused on finding semantic correspondences~\cite{ufer2017deep,rocco2018end} across different objects of the same category. Beside of visual information, a recent study~\cite{yi2018learning} explores to encode motion context for identifying outliers from keypoint matches, i.e., 4-d coordinate pairs, while we aim to exploit geometric context from single image without any reference. Overall, encoding proper context is non-trivial and still unclear in 2D local feature learning.

\begin{figure*}[th]
	\centering 
	\includegraphics[width=\textwidth]{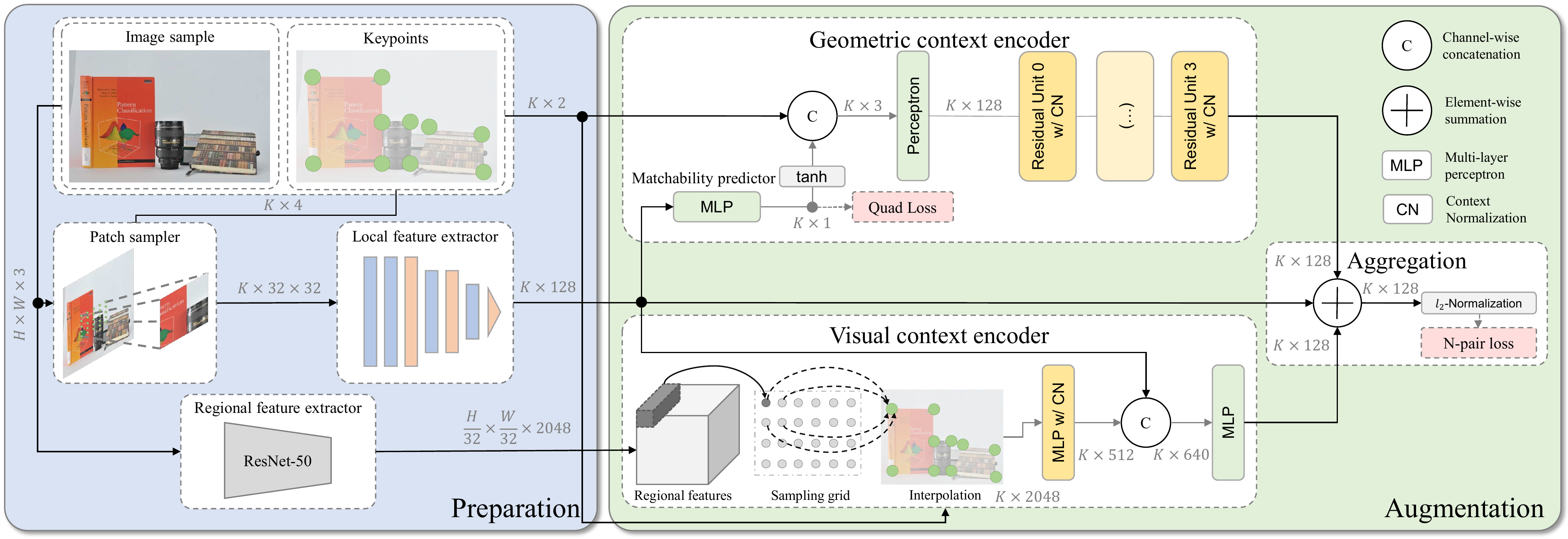}
	\caption{The proposed augmentation framework consumes a single image as input, from which 2D keypoints, local and regional features are extracted and encoded as geometric and visual context to improve the raw local feature description.}
	\label{fig:arch}
\end{figure*}

\smallskip\noindent\textbf{Point feature learning.}
In the present work, one of our goals is to explore geometric features from keypoint distribution, we thus resort to PointNet~\cite{qi2017pointnet} and its variants~\cite{qi2017pointnet++,deng2018ppfnet,yi2018learning} to consume unordered points. Although great success has been witnessed in learning tasks on 3D points, there are only few studies exploiting the potential outcome of 2D keypoint sets. In essence, keypoint structure is not intuitively meaningful and robust, as being highly dependent on the performance of interest point detectors and strongly affected by image variations. However, in descriptor learning, we consider the keypoint location as an important cue that bridges each individual local feature that has potentials to alleviate the local visual ambiguity.

\smallskip\noindent\textbf{Loss formulation.}
Recent local descriptors are often evolved with advanced variants of N-pair losses. Initially, L2-Net~\cite{tian2017l2} adopts a log-likelihood formulation, which is later extended by HardNet~\cite{mishchuk2017working} with a subtractive hinge loss. Furthermore, GeoDesc~\cite{Luo_2018_ECCV} applies an adaptive margin to improve the convergence in terms of different hard negative mining strategies, while AffNet~\cite{AffNet2017} approaches the same issue by fixing the distance to hardest negative sample during training. Meanwhile, on the other hand, DOAP~\cite{he2018local} extends the N-pair loss to a list-wise ranking loss, while~\cite{keller2018learning} points out and studies the scale effects in N-pair losses while introducing additional manual tuning of hyper-parameters. Principally, a good loss is supposed to encourage similar patches to be close while dissimilar ones to be distant in the descriptor space. In this spirit, we aim to further resolve the scale effects in~\cite{keller2018learning} in an self-adaptive manner, without the need of complex heuristics or manual tuning.

\section{Local Descriptor Augmentation}
\smallskip\noindent\textbf{Overview.} As illustrated in Fig.~\ref{fig:arch}, the proposed framework consists of two main modules: \emph{preparation} (left) and \emph{augmentation} (right). The \emph{preparation} module provides input signals in different modalities (raw local feature, high-level visual feature and keypoint location), which are then fed to the \emph{augmentation} module and aggregated into compact feature descriptions.  At test time, the augmentation needs to be performed once per image, resulting in $K$ feature vectors for $K$ corresponding keypoints.
\subsection{Preparation}
\label{sec:prep}
\smallskip\noindent\textbf{Patch sampler.}
This module takes images and their keypoints as input, producing $32\times 32$ gray-scale patches. Akin to~\cite{yi2016lift,Luo_2018_ECCV}, image patches are sampled by a spatial transformer~\cite{jaderberg2015spatial}, whose parameters are derived from keypoint attributes (coordinates, orientation and scale) from the SIFT detector. As a result, the sampled patch has the same support region size with the SIFT descriptor.

\smallskip\noindent\textbf{Local feature extractor.}
This module takes image patches as input, producing 128-d feature descriptions as output. We borrow the lightweight 7-layer convolutional networks as used in several recent works~\cite{tian2017l2,mishchuk2017working,Luo_2018_ECCV}.

\smallskip\noindent\textbf{Regional feature extractor.}
In contrast to aggregating features of different domain sizes~\cite{zagoruyko2015learning,kumar2016learning,tian2017l2}, in the present work, we fix the sampling scale of patches, and exploit contextual cues by inspiration of well-studied regional representation in image retrieval tasks~\cite{tolias2015particular,radenovic2016cnn,noh2017large}. Without the loss of generality, we reuse features from an off-the-shelf deep image retrieval model of ResNet-50~\cite{he2016deep}. As in~\cite{tolias2015particular}, feature maps are extracted from the last bottleneck block, across which each response is regarded as a regional feature vector effectively corresponding to a particular region in the image. As a result, we derive regional features of $\frac{H}{32}\times\frac{W}{32}\times 2048$, where $H$ and $W$ denote the original image height and width. The aggregation of regional and local features will be later discussed in Sec.~\ref{sec:vc}.

\subsection{Geometric context encoder}
\label{sec:gc}

This module takes $K$ unordered points as input, and outputs 128-d corresponding feature vectors. Each input point is represented as 2D keypoint coordinate, and can be associated with other attributes. 

\smallskip\noindent\textbf{2D point processing.}
At first glance, 2D keypoints are inappropriate to serve as robust contextual cues, as its presence is heavily dependent on image appearance and thus affected by various image variations. As a result, keypoint distribution depicting the same scene may suffer from significant density or structure variations, as examples shown in Fig.~\ref{fig:geo_structure_example}. Hence, acquiring strong invariance property is the key challenge when designing the context encoder.

Initially, we attempt to approach the goal by PointNet~\cite{qi2017pointnet} and its variants~\cite{qi2017pointnet++,deng2018ppfnet}. Although having shown great success on processing 3D point clouds, those prevalent PointNet methods fails to achieve consistent improvement in terms of 2D points processing (Sec.~\ref{sec:ablation_context}). Instead, we resort to~\cite{yi2018learning}, in which context normalization (CN) is equipped in PointNet and consumes putative matches ($4$-d coordinate pairs) for outlier rejection in image matching. In this work, we aim to further explore the usability of CN for modeling 2D point distribution in single image.

\begin{figure}[h]
	\centering 
	\includegraphics[width=0.48\textwidth]{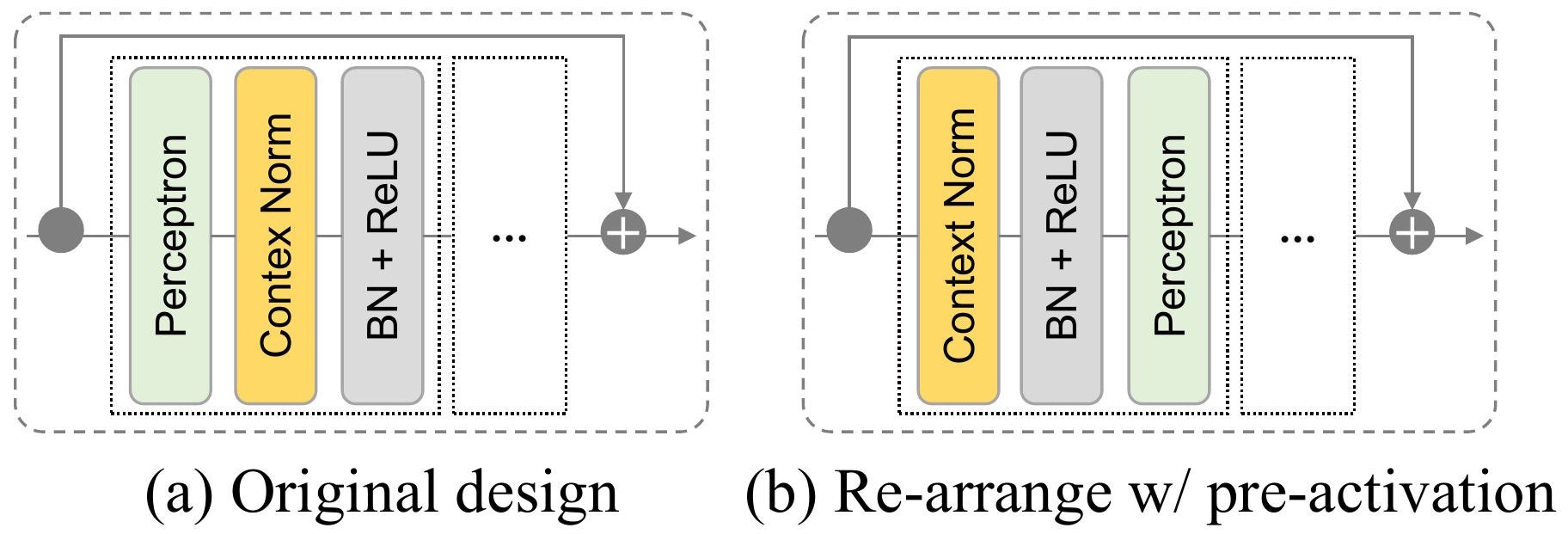}
	\vspace{-6mm}
	\caption{Different designs of residual unit with context normalization, where re-arranging with \emph{pre-activation} improves by a considerable margin than its counterpart.}
	\label{fig:context_norm}
\end{figure}

Formally, CN is a non-parametric operation that simply normalizes feature maps according to their distribution, written as
$\hat{\bm{o}}_i^l = \frac{(\bm{o}_i^l-\bm{\mu}^l )}{\bm{\sigma}^l}$,
where $\bm{o}_i^l$ is the output of $i$-th point in layer $l$, and $\bm{\mu}^l, \bm{\sigma}^l$ are mean and standard deviation of the output in layer $l$. To equip the operation, we borrow the residual architecture in~\cite{yi2018learning}, where each residual unit is built with perceptrons followed by context and batch normalization, as illustrated in Fig.~\ref{fig:context_norm}{\color{red}a}. 

However, the above design leads to a \emph{non-negative} output from the residual branch that may impact the representational ability as investigated in~\cite{he2016identity} and witnessed in our experiments (Sec.~\ref{sec:ablation_context}). Following the teachings of~\cite{he2016identity}, we re-arrange the operations in each residual unit with \emph{pre-activation}, which is compatible with CN as presented in Fig.~\ref{fig:context_norm}{\color{red}b}. We then construct four such units for the encoder, as shown in Fig.~\ref{fig:arch}. We will show that this simple revision plays an important role to ease the optimization. 

Intuitively, the non-parametric CN suffices to model the keypoint distribution in our task, while high-level abstractions (e.g., in PointNet++~\cite{qi2017pointnet++}) may not be necessary.

\smallskip\noindent\textbf{Matchability predictor.}
In 3D point cloud processing, low-level color and normal~\cite{qi2017pointnet} information or complex geometric attributes~\cite{deng2018ppfnet} are often incorporated to enhance the representation. Similarly, associating 2D coordinate input with other meaningful attributes would be promising to boost the performance. However, due to the substantial variations, e.g. perspective change, it is non-trivial to define appropriate intermediate attributes on 2D points.

Although this issue has been merely discussed, we draw inspiration from~\cite{hartmann2014predicting}, which poses a problem named \emph{matchability prediction} that targets to decide \emph{whether a keypoint descriptor is matchable before the matching stage}. In practice, the matchability serves as learned attenuation to diversify the keypoints, so that the feature encoder can implicitly focus on the points that are more robust, i.e., matchable, in order to improve the invariance property.

In the present work, we approach the matchability prediction with deep learning techniques instead of a random forest in~\cite{hartmann2014predicting}, and constrain the prediction to be consistent between images. Inspired by learning-based keypoint detection methods~\cite{savinov2017quad,zhang2018learning}, we resort to an unsupervised learning scheme that aims to appropriately rank points by their matchability. Formally, given $K$ correspondences $(p_1^n, p_2^n)$, $n\in[1, K]$ from an image pair, we first extract their local features $(\bm{f}_1^n, \bm{f}_2^n)$, then construct \emph{feature quadruples} as $(\bm{f}_1^i, \bm{f}_1^j, \bm{f}_2^i, \bm{f}_2^j)$, satisfying $i, j \in [1, K], i \neq j$ and holding that:
\begin{equation}  
\label{eq:cond}
\left\{
\begin{array}{rcl}
H(\bm{f}_1^i) > H(\bm{f}_1^j) & \& & H(\bm{f}_2^i) > H(\bm{f}_2^j) \\
&   \text{or}   & \\
H(\bm{f}_1^i) < H(\bm{f}_1^j) & \& & H(\bm{f}_2^i) < H(\bm{f}_2^j) 
\end{array} \right.,
\end{equation} 
where $H(:)$ absorbs the raw local feature into a single real-valued matchability, implemented as standard multi-layer perceptrons (MLPs). Here, Cond.~\ref{eq:cond} aims to preserve a ranking of each keypoint, hence improves the repeatability of prediction. The condition can be re-written as:
\begin{equation}
\begin{aligned}
& R(\bm{f}_1^i, \bm{f}_1^j, \bm{f}_2^i, \bm{f}_2^j) = \\ 
& (H(\bm{f}_1^i) - H(\bm{f}_1^j))(H(\bm{f}_2^i) - H(\bm{f}_2^j)) > 0,
\end{aligned}
\end{equation}
the final objective can be obtained with a hinge loss:
\begin{equation}
\label{eq:quad_loss}
\mathcal{L}_{quad} = \frac{1}{K(K-1)}\sum_{i, j, i\neq j}\max(0, 1 - R(\bm{f}_1^i, \bm{f}_1^j, \bm{f}_2^i, \bm{f}_2^j)).
\end{equation}

\begin{figure}[h]
	\centering 
	\includegraphics[width=0.38\textwidth]{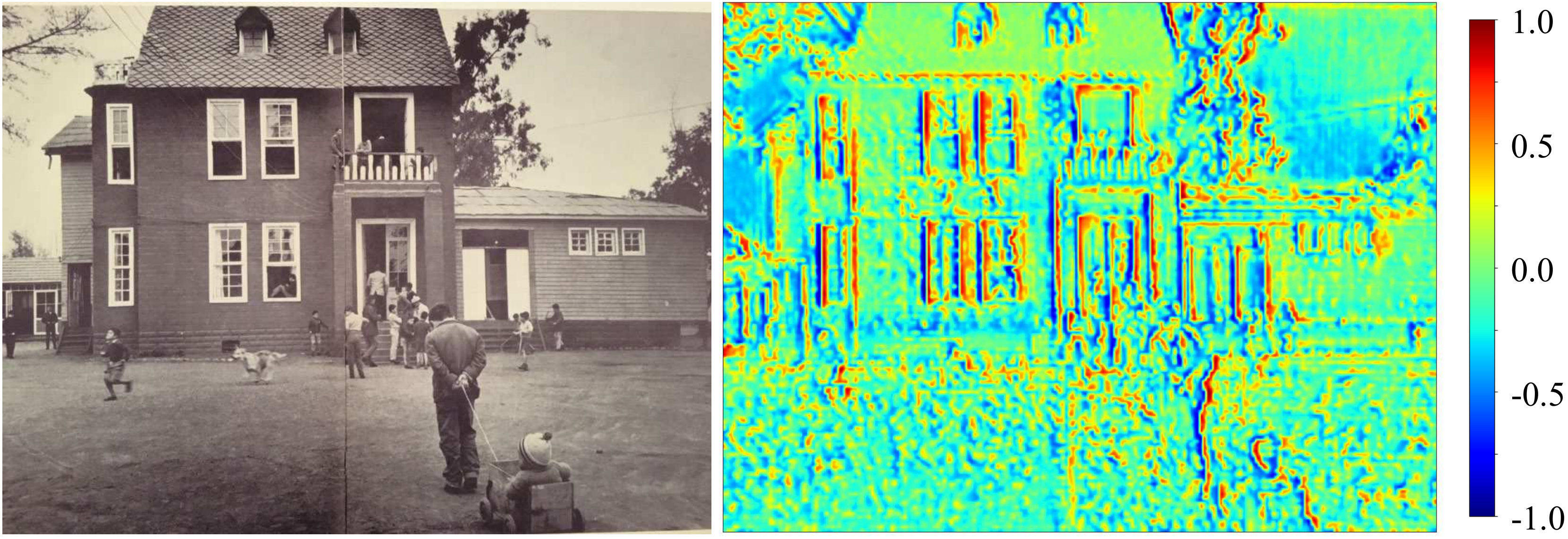}
	\vspace{-2mm}
	\caption{Visualization of matchability responding to the entire image (best viewed in color).}
	\label{fig:matchability}
	\vspace{-4mm}
\end{figure}

In the proposed framework, the matchability is learned as an auxiliary task, which is then activated by $\tanh$ and associated with keypoint coordinates as the network input, as in Fig.~\ref{fig:arch}. Beside of Eq.~\ref{eq:quad_loss}, the gradient from final augmented features will flow through the matchability predictor, allowing a joint optimization of the entire encoder. The visualization of predicted matchability is shown in Fig.~\ref{fig:matchability}.

\subsection{Visual context encoder}
\label{sec:vc}
This module consumes regional features of $\frac{H}{32}\times\frac{W}{32}\times 2048$ in Sec.~\ref{sec:prep}, $K$ local features and their location, and  produces $K$ augmented features. To integrate visual information in different levels, a valid option as in~\cite{deng2018ppfnet} is to concatenate the global representation of entire image on raw local features. In our framework, the global feature can be derived by applying Maximum Activations of Convolutions (MAC) aggregation~\cite{radenovic2016cnn}, which simply max-pools over all dimensions of regional features. However, such compact representation is shown to obscure the raw local description, due to the lack of spatial distinctions (Sec.~\ref{sec:ablation_context}). Hence, we stick to the regional representation, where the key issue is to handle the regional features and keypoints of different numbers ($\frac{H}{32}\times\frac{W}{32}$ and $K$). 

To achieve the goal, we associate regional features to a regular sampling grid on the image, then interpolate $\frac{H}{32}\times\frac{W}{32}$ grid points at coordinates of the $K$ keypoints. For interpolation, we use inverse distance weighted average based on $k$ nearest neighbors (in default we use $k=3$), formulated as:
\begin{equation}
\bm{f}(\hat{p}_i)=\frac{\sum_{j=1}^k w(p_j)\bm{f}(p_j)}{\sum_{j=1}^k w(p_j)}, \text{and}~~w(p_j) = \frac{1}{d(\hat{p}_i,p_j)},
\end{equation}
where $\bm{f}(:)$ is the regional feature located at a certain grid point. $\hat{p}_i$, $i\in[1, N]$ and $p_j$, $j\in[1, \frac{H}{32}\times\frac{W}{32}]$ indicate the interpolated and original grid point. Next, the dimensionality is reduced by applying point-wise MLPs, where we also insert  CN after each perceptron in order to capture global context. Finally, raw local features are concatenated and further mapped by MLPs, forming the final 128-d features. The above process is illustrated in Fig.~\ref{fig:arch}.

\subsection{Feature aggregation with raw local feature}
\label{sec:agg}
To aggregate the above two types of contextual features, similar to the CS structure, one option is to concatenate them together and forms features of, in our case, $384$-d ($128\times3$). However, the increased dimensionality will introduce excessive computational cost in the matching stage of $\mathcal{O}(n^2)$ complexity. Instead, as shown in Tab.~\ref{fig:arch}, we propose to combine different feature streams into a single vector by element-wise summation and L2-normalization, i.e., without altering the feature dimensionality. Beside of the simplicity, such strategy allows flexible use of the proposed augmentation. For example, in situations where regional features are not available, one may aggregate with only geometric context without the need of retraining the model.

\subsection{N-pair loss with softmax temperature}
\label{sec:loss}
N-pair losses have been primarily used by recent works. Empirically, the subtractive hinge loss~\cite{mishchuk2017working,Luo_2018_ECCV,AffNet2017} has reported better performance, of which the main idea is to push similar samples away from dissimilar ones to a certain \emph{margin} in the descriptor space. However, setting the appropriate margin is tricky, which does not always assure convergence as observed in~\cite{Luo_2018_ECCV,AffNet2017}. More generally, the criteria of making a good loss is studied in~\cite{keller2018learning}, from which guidelines are provided on tuning loss parameters on a particular dataset. In this spirit, we aim to further ease the pain of parameter search in~\cite{keller2018learning}, and obtain an adaptive loss that allows fast convergence regardless of the learning difficulty. 

We use the log-likelihood form of N-pair loss~\cite{tian2017l2} as a base, which originally does not involve any tunable parameter. Formally, given L2-normalized feature descriptors $\mathbf{F}_1=[\bm{f}_1^1 \bm{f}_1^2 ... \bm{f}_1^N]^T, \mathbf{F}_2=[\bm{f}_2^1 \bm{f}_2^2 ... \bm{f}_2^N]^T\in{\mathbb{R}^{N\times 128}}$, the distance matrix $\mathbf{D}=[d_{ij}]_{N\times N}$ can be obtained by $\mathbf{D}=\sqrt{2(1-\mathbf{F}_1\mathbf{F}_2^T)}$. By applying both row-wise ($r$) and column-wise ($c$) softmax, we derive the final loss as:
\begin{equation}
\label{eq:npair}
\begin{aligned}
& \mathcal{L}_{N\text{-}pair}=-\frac{1}{2}(\sum_{i}\log{s_{ii}^r} + \sum_{i}\log{s_{ii}^c}), \\
& \text{where}~~[s_{ij}]_{N\times N}=\text{softmax}(2 - \mathbf{D}).
\end{aligned}
\end{equation}

Noted that since input features are L2-normalized, the resulting $d_{ij}$ is bounded by $[0, 2]$, which causes convergence issues due to the scale sensitivity of softmax function~\cite{hoffer2018fix}. Similarly, we introduce a single trainable parameter $\alpha$, referred to as \emph{softmax temperature}, to amend the inability of re-scaling the input. The loss now becomes:
\begin{equation}
[s_{ij}]_{N\times N} = \text{softmax}(\alpha(2 - \mathbf{D})),
\end{equation}
where $\alpha$ is initialized to $1$ and regularized with the same
weight decay in the network, hence does not require any manual tuning or complex heuristics. In the experiments in Sec.~\ref{sec:ablation_st}, we show this simple technique improves drastically than its original form~\cite{tian2017l2}, whose performance we suspect is hindered due to the above-mentioned scale sensitivity.  In the proposed framework, we compute the N-pair loss on augmented features, and obtain the total loss:
\begin{equation}
\label{equ:tot_loss}
\mathcal{L}_{total}= \mathcal{L}_{N\text{-}pair} + \lambda\mathcal{L}_{quad},
\end{equation}
where we choose $\lambda=1$ in the experiment.

\section{Experiments}
\subsection{Implementation}
\label{sec:implementation}

\smallskip\noindent\textbf{Training details.}
Although the framework is end-to-end trainable, we \emph{fix} the local and regional feature extractors in Sec.~\ref{sec:prep} during the training, in order to clearly demonstrate the efficacy of the proposed augmentation scheme. We train the networks using SGD with a base learning rate of 0.05, weight
decay of 0.0001 and momentum at 0.9. The learning rate exponentially decays by 0.1 for every 100k steps. The batch size is set to 2, and each time 1024 keypoints are randomly sampled including random numbers of matchable and noisy keypoints (see Appendix~{A.\color{red}1}). Input patches are standardized to have zero mean and unit norm, while input keypoint coordinates are normalized to $[-1, 1]$ regarding the image size.

\smallskip\noindent\textbf{Training dataset.} Although UBC Phototour~\cite{brown2007automatic} is used as a common practice, this dataset consists of only three scenes with limited diversity of keypoint distribution. In order to achieve better generalization ability, we resort to large-scale photo-tourism~\cite{wilson2014robust,radenovic2016cnn} and aerial datasets (GL3D)~\cite{shen2018matchable} as in~\cite{yi2016lift,Luo_2018_ECCV}, and generate groundtruth matches from SfM. We manually exclude the data that is used in the evaluation.

\smallskip\noindent\textbf{Data augmentation.}
We randomly perturb input patches by affine transformations including rotation (90$\degree$), anisotropic scaling and translation w.r.t. the detection scale. For keypoint augmentation, we perturb the coordinate with random homography transformation as in~\cite{detone2016deep} (see Appendix~{A.\color{red}1}).
\subsection{Evaluation datasets}
\smallskip\noindent\textbf{Homography dataset.}
HPatches~\cite{balntas2017hpatches} is a large-scale patch dataset for evaluating local features regarding illumination and viewpoint changes. As groundtruth homographies and raw images are provided, HPatches can also be used to evaluate image matching performance, which we accordingly refer to as HPSequences as in~\cite{lenc2018large}, consisting of 116 sequences and 580 image pairs.

\smallskip\noindent\textbf{Wild dataset.} Similar to settings in~\cite{yi2018learning}, we also evaluate on outdoor YFCC100M~\cite{thomee2015yfcc100m} (1000 pairs) and indoor SUN3D~\cite{xiao2013sun3d} (539 pairs) datasets. Compared with HPSequences, the two datasets additionally introduce variations such as self-occlusions, and in particular, repetitive or feature-poor patterns in indoor scenes, which is generally considered challenging for sparse matching.

\smallskip\noindent\textbf{SfM dataset.} Following~\cite{schonberger2017comparative}, we evaluate on SfM dataset such as well-known \emph{Fountain} and \emph{Herzjesu}~\cite{strecha2008benchmarking}, or landmark collections~\cite{wilson2014robust}. We integrate the proposed framework into SfM pipeline, i.e., COLMAP~\cite{schoenberger2016sfm}, and use the  keypoints provided in~\cite{schonberger2017comparative} to compute the local features.

\begin{table*}[ht]
	\centering
	\caption{Comparisons on HPSequences~\cite{balntas2017hpatches} of different designs of visual and geometric context encoder, and the performance of entire augmentation scheme. `i/v' denotes two evaluations on \emph{illumination} and \emph{viewpoint} sequences, respectively.}
	\vspace{-4mm}
	\resizebox{\textwidth}{!}{  
		\begin{tabular}{l|cc|l|cc|l|cc}
			\Xhline{1pt}
			\multicolumn{3}{c|}{\textbf{Visual context encoder}}                            & \multicolumn{3}{c|}{\textbf{Geometric context encoder}}                                     & \multicolumn{3}{c}{\textbf{Comparison with other methods}}                    \\ \hline
			\multicolumn{1}{c}{\textit{Strategy}} & \multicolumn{2}{c}{\textit{Recall i/v}} & \multicolumn{1}{c}{\textit{Network architecture}} & \multicolumn{2}{c}{\textit{Recall i/v}} & \multicolumn{1}{c}{\textit{Method}} & \multicolumn{2}{c}{\textit{Recall i/v}} \\ \hline
			baseline (GeoDesc~\cite{Luo_2018_ECCV})                       & 59.46                 & 71.24                 &               baseline (GeoDesc~\cite{Luo_2018_ECCV})                   &          59.46        &        71.24          & SIFT~\cite{lowe2004distinctive}                                &       47.36           &       53.06           \\
			CS (256-d)~\cite{zagoruyko2015learning,kumar2016learning,tian2017l2} & 59.83                 & 71.27                  & PointNet~\cite{qi2017pointnet}                            &       59.61           &     70.96             & L2-Net~\cite{tian2017l2}                             & 47.58                 & 53.96                 \\
			w/ global feature~\cite{deng2018ppfnet}                         &      59.11            &    71.02              & w/ CN (pre.) + xy                             & 61.67                 & 72.63                & HardNet~\cite{mishchuk2017working}                             & 57.63                 & 63.36                 \\
			w/ regional feature                                 & 63.64                 & 73.37                 &            w/ CN (pre.) + xy + raw local feature                 & 60.91                 & 72.99                 & GeoDesc~\cite{Luo_2018_ECCV}                             & 59.46                 & 71.24                 \\
			\textbf{w/ regional feature + CN}                     & \textbf{63.98}                 & \textbf{73.63} &      w/ CN (orig.) + xy + matchability               &     59.94       &     71.25           & \textbf{ContextDesc} & \textbf{66.55} &   \textbf{75.52}  \\
			& &                &      \textbf{w/ CN (pre.) + xy + matchability}               &       \textbf{62.82}           &       \textbf{73.40}  & \textbf{ContextDesc+} & \textbf{67.14} & \textbf{76.42} \\ \Xhline{1pt} 
		\end{tabular}
		\label{tab:ablation}
	}
	\vspace{-2mm}
\end{table*}

\subsection{Evaluation protocols}
\label{sec:metric}

\smallskip\noindent\textbf{Patch level.}
For HPatches~\cite{balntas2017hpatches}, we follow its evaluation protocols and use mean average precision (mAP) for three subtasks, including patch verification, matching, and retrieval.

\smallskip\noindent\textbf{Image level.}
For HPSequences, we use \emph{Recall = \# Correct Matches / \# Correspondences} defined in~\cite{heinly2012comparative}, to quantify the image matching performance, where \emph{\# Correct matches} are matches found by nearest neighbor searching and verified by groundtruth geometry, e.g., homography, while \emph{\# Correspondences} are matches that should have been identified by the given keypoint
locations. Following~\cite{heinly2012comparative}, a match point is determined to be correct if it is within 2.5 pixels from the wrapped keypoint in the reference image. We use a standard SIFT detector to localize the keypoints, of which the number is randomly sampled to 2048. For YFCC100M~\cite{thomee2015yfcc100m} and SUN3D~\cite{xiao2013sun3d}, we follow the same setting in~\cite{yi2018learning} and report the median number of inlier matches after RANSAC for each dataset. 

\smallskip\noindent\textbf{Reconstruction level.} For clarity, we report metrics in~\cite{schonberger2017comparative} that quantify the completeness of SfM, including the number of registered images (\emph{\# Registered}), sparse points (\emph{\# Sparse Points}) and image observations (\emph{\# Observations}).

\subsection{Ablation study}
\subsubsection{Design of context encoder}
\label{sec:ablation_context}
In this section, we evaluate two splits of HPSequences~\cite{balntas2017hpatches}: \emph{illumination (i)} and \emph{viewpoint (v)}, regarding different image transformations. We report \emph{Recall} as defined in Sec.~\ref{sec:metric}. 
If not specified, we use GeoDesc~\cite{Luo_2018_ECCV} as a baseline model (\emph{baseline (GeoDesc)}) to extract raw local features, whose parameters are \emph{fixed} during the training of augmentation.

\smallskip\noindent\textbf{Visual context.} We compare four designs, including i) \emph{CS (256-d)}: the central-surround (CS) structure~\cite{zagoruyko2015learning,kumar2016learning,tian2017l2} as described in Sec.~\ref{sec:related_work}, which concatenates local features from different domain sizes. ii) \emph{w/ global feature}: the integration with global  features~\cite{deng2018ppfnet}, which is originally designed for improving 3D local descriptors. iii) \emph{w/ regional feature}: the proposed integration with interpolated regional features, and its variant iv) \emph{w/ regional feature + CN}: with context normalization to incorporate global visual information.	

As shown in Tab.~\ref{tab:ablation} (left columns), the CS structure~\cite{zagoruyko2015learning,kumar2016learning,tian2017l2} delivers only marginal improvements despite the doubled dimensionality. Meanwhile, though being effective in 3D descriptor learning, the integration with global features~\cite{deng2018ppfnet} instead harms the performance, which we ascribe to the limited representation ability of a single global feature. Finally, the proposed integration with interpolated regional features shows clear improvements, as it better handles both spatial and visual distinctiveness. Moreover, to strengthen global context awareness, we show that the performance can be further boosted by equipping context normalization when encoding regional features.

\smallskip\noindent\textbf{Geometric context.} We study five options: i) PointNet-like architecture, i.e., segmentation networks in~\cite{qi2017pointnet} without the final classifier. ii) Pre-activated context normalization (CN) networks in Sec.~\ref{sec:gc} with 2D xy input, and its variants iii) with additional raw local feature input or iv) with matchability. We also compare the use of pre-activation of the residual unit in context normalization networks.

As presented in Tab.~\ref{tab:ablation} (middle columns), though widely used in processing 3D points, PointNet~\cite{qi2017pointnet} does not perform well in our task, while the similar phenomenon is also observed in~\cite{yi2018learning} when processing 2D correspondences. Besides, it is noticed that input with raw local feature does not help to boost the performance, which we attribute to the weak relevance between local features as extracted from different orientations and levels of scale space pyramid. Instead, the incorporation with matchability is notably beneficial, as matchability is more comprehensive as a high-level abstraction of local feature. Finally, the pre-activation is clearly a preferable alternative than its original design. 

\smallskip\noindent\textbf{Integration with cross-modality context.}
Finally, we evaluate the full augmentation with both visual and geometric context (\emph{ContextDesc}). As shown in Tab.~\ref{tab:ablation} (right columns), the simple summation aggregation in Sec.~\ref{sec:agg} effectively takes advantage of both context, delivering remarkable improvements over the state-of the-art.

\subsubsection{Efficacy of softmax temperature in N-pair loss}
\label{sec:ablation_st}

To demonstrate the validity of proposed loss in Sec.~\ref{sec:loss}, we train \emph{only} the local base model without any context awareness, and compare different losses including: i) the plain N-pair loss in~\cite{tian2017l2} without scale temperature, and ii) the scale-aware loss in~\cite{keller2018learning} with its original parameters.

\begin{table}[ht]
	\centering
	\caption{Evaluation results on 1) HPatches~\cite{balntas2017hpatches} of three complementary tasks: patch verification, matching and retrieval. 2) HPSequences of two sequence splits.}
	\label{tab:hpatches_eval}
	\resizebox{0.48\textwidth}{!}{  
		\begin{tabular}{cc|ccc}
			\Xhline{1pt}
			& \textbf{GeoDesc~\cite{Luo_2018_ECCV}} & \textbf{w/ loss in~\cite{tian2017l2}}  & \textbf{w/ loss in~\cite{keller2018learning}} &  \textbf{Ours} \\ \Xhline{1.0pt}
			\multicolumn{5}{c}{\emph{HPatches, mAP [\%]}} \\ \Xhline{0.7pt}
			\emph{Verification} & \textbf{91.1} & 78.3 & 81.2 & 90.2  \\
			\emph{Matching} & 59.1 & 23.9 & 40.5 & \textbf{59.2}  \\
			\emph{Retrieval} & 74.9 & 46.8 & 64.0 & \textbf{76.0}  \\
			 \Xhline{0.7pt}
			\multicolumn{5}{c}{\emph{HPSequences, Recall}} \\ 
			\Xhline{0.7pt}
			\emph{Seq. i} & 59.5 & 32.2 & 50.0 & \textbf{59.7}  \\
			\emph{Seq. v} & 71.2 & 48.5 & 64.8 & \textbf{72.6}  \\
			\Xhline{1pt}
		\end{tabular}
	}

\end{table}

\begin{figure*}[t]
	\centering 	
	\includegraphics[width=\textwidth]{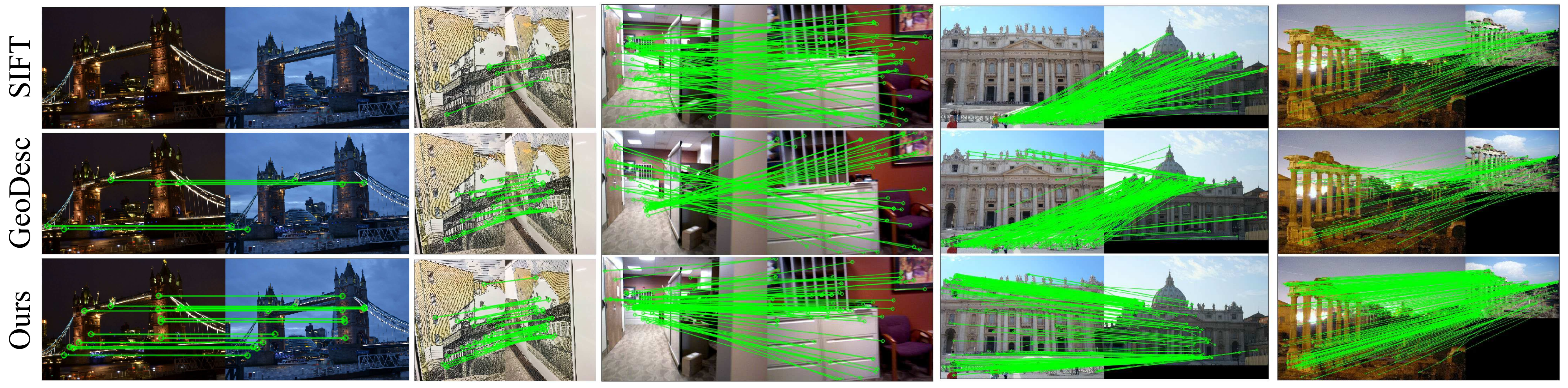}
	\vspace{-5mm}
	\caption{Matching results after RANSAC in different challenging scenarios.  From top to bottom: SIFT, GeoDesc and ours.  The augmented feature helps to find more inlier matches, and further allows a more accurate recovery of camera geometry.}
	\label{fig:vis}
	\vspace{-1.5mm}
\end{figure*}

As shown in Tab.~\ref{tab:hpatches_eval}, the proposed loss improves the overall performance over the previous best-performing GeoDesc~\cite{Luo_2018_ECCV} under similar training settings, while GeoDesc requires additional geometric supervision. Besides, the proposed loss clearly shows better convergence compared with losses in~\cite{tian2017l2} and~\cite{keller2018learning}. Although we suspect that the loss in~\cite{keller2018learning} may perform better with careful parameter searching, the proposed loss is advantageous due to its self-adaptivity without the need of complex heuristics or manual tuning.

Moreover, once replace GeoDesc with the above model as a base in the augmentation scheme, the final performance can be further improved by a significant margin, denoted as \emph{ContextDesc+} in Tab.~\ref{tab:ablation} (right columns), which again addresses the advance of improved base model. We will use this model to complete the following experiments.

\subsection{Generalization}
\label{sec:generalization}

\smallskip\noindent\textbf{Wild dataset.} The evaluation results on two challenging datasets (\emph{outdoor} YFCC100M~\cite{thomee2015yfcc100m} and \emph{indoor} SUN3D~\cite{xiao2013sun3d}) are presented in Tab.~\ref{tab:wild}. The proposed cross-modality context augmentation delivers $\scriptsize{\sim}35\%$ and $\scriptsize{\sim}125\%$ improvements over the previous state of the art, which effectively demonstrates the strong generalization ability of the learned augmented features in practical scenes.

\begin{table}[h]
	\centering
	\caption{Evaluation results on wild datasets: \emph{indoor} SUN3D~\cite{xiao2013sun3d} and \emph{outdoor} YFCC100M~\cite{thomee2015yfcc100m} datasets.}
	\resizebox{0.48\textwidth}{!}{\begin{tabular}{cccccc}
			\Xhline{1pt}
			& \textbf{SIFT~\cite{lowe2004distinctive}} & \textbf{L2-Net~\cite{tian2017l2}} & \textbf{HardNet~\cite{mishchuk2017working}} & \textbf{GeoDesc~\cite{Luo_2018_ECCV}} & \textbf{Ours} \\ \hline
			\multicolumn{6}{c}{\textit{median number of inlier matches}}                                   \\ \hline
			\textit{indoor}  &   138          &       153          &        239          &       271        & \textbf{365}                    \\
			\textit{outdoor} &    168           &         173        &         219         &         214         &      \textbf{482}                  \\ \Xhline{1pt}
		\end{tabular}
		\label{tab:wild}
	}
\end{table}

\smallskip\noindent\textbf{SfM dataset.} We further demonstrate the improvement in complex SfM pipeline. As shown in Tab.~\ref{tab:sfm}, the integration of augmented feature generalizes well among different scenes even in large-scale SfM tasks, meanwhile consistently boosts the completeness of sparse reconstruction. Some matching results are presented in Fig.~\ref{fig:vis}, and more visualizations can be found in the appendix.

\begin{table}[th]
	\centering
	\caption{Evaluation results on SfM dataset~\cite{schonberger2017comparative}.}
	\label{tab:sfm}
	\resizebox{0.48\textwidth}{!}{  
		\begin{tabular}{cccccccc}
			\Xhline{1pt}
			& & \textbf{\# Images} & \textbf{\# Registered} & \textbf{\# Sparse Points} & \textbf{\# Observations} \\ \Xhline{0.7pt}  
			
			\textbf{Fountain} & \emph{SIFT~\cite{lowe2004distinctive}} & 11 & 11 & 10,004 & 44K \\ 
			& \emph{GeoDesc~\cite{Luo_2018_ECCV}} &  & 11 & 16,687 & 83K \\ 
			& \emph{Ours} &  & 11 & \textbf{16,965} & \textbf{84K} \\ \Xhline{0.7pt} 
			
			\textbf{Herzjesu} & \emph{SIFT} & 8 & 8 & 4,916 & 19K & \\ 
			& \emph{GeoDesc} &  & 8 & 8,720 & 38K \\ 
			& \emph{Ours} &  & 8 & \textbf{9,429} & \textbf{40K} \\ \Xhline{0.7pt} 
			
			\textbf{South Building} & \emph{SIFT} & 128 & 128 & 62,780 & 353K \\ 
			& \emph{GeoDesc} & & 128 & 170,306 & 887K \\ 
			& \emph{Ours} & & 128 & \textbf{174,359} & \textbf{893K} \\ \Xhline{0.7pt} 
			

			\textbf{Roman Forum} & \emph{SIFT} & 2,364 & 1,407 & 242,192 & 1,805K \\ 
			& \emph{GeoDesc} & & 1,566 & 770,363 & 5,051K \\ 
			& \emph{Ours} & & \textbf{1,571} & \textbf{848,319} & \textbf{5,484K}\\
			\Xhline{0.7pt}
			\textbf{Alamo} & \emph{SIFT} & 2,915 & 743 & 120,713 & 1,384K \\
			& \emph{GeoDesc} & & 893 & 353,329 & 3,159K \\
			& \emph{Ours} & & \textbf{921} & \textbf{424,348} & \textbf{3,488K}\\ 
			\Xhline{1.0pt}
		\end{tabular}
	}
\vspace{-2mm}
\end{table}

\subsection{Discussions on practicability}

\smallskip\noindent\textbf{Invariance property.}
We again use \emph{Recall} and evaluate on Heinly benchmark~\cite{heinly2012comparative} to quantify the invariance property. As shown in Tab~\ref{tab:heinly}, the proposed method improves remarkably over the previous best-performing descriptor, except for some minor underperformance regarding \emph{Rotation} change when images are rotated up to 180\textdegree, which may be caused by the inability of being fully rotation-invariant especially for the regional feature extractor.

\begin{table}[th]
	\centering
	\caption{Evaluation results regrading different transformations on Heinly benchmark~\cite{heinly2012comparative}.}
	\resizebox{0.32\textwidth}{!}{\begin{tabular}{lccc}
		\Xhline{1pt}
		& \textbf{SIFT~\cite{lowe2004distinctive}} & \textbf{GeoDesc~\cite{Luo_2018_ECCV}} & \textbf{Ours} \\ \hline
		\multicolumn{4}{c}{\textit{Recall}} \\ \hline 
		\textit{JPEG}           & 60.7          & 66.1             & \textbf{78.6}           \\
		\textit{Blur}           & 41.0          & 47.7             & \textbf{57.8}           \\
		\textit{Exposure}       & 78.2          & 86.4             & \textbf{88.2}           \\
		\textit{Day-Night}      & 29.2          & \textit{39.6}    & \textbf{43.3}           \\
		\textit{Scale}          & 81.2          & 85.8             & \textbf{88.1}           \\
		\textit{Rotation}       & 82.4          & \textbf{87.6}             & 86.3           \\
		\textit{Scale-Rotation} & 29.6          & 33.7             & \textbf{38.0}           \\
		\textit{Planar}         & 48.2          & 59.1             & \textbf{61.7}           \\ \Xhline{1pt}
	\end{tabular}
}
	\label{tab:heinly}
	\vspace{-1mm}
\end{table}

\smallskip\noindent\textbf{Computational cost.} 
Towards practicability, we only use shallow MLPs or non-parametric context normalization in the augmentation framework, which thus introduces only insignificant computation overhead. As reported in Tab.~\ref{tab:cost}, suppose that regional features are readily extracted, e.g., from a retrieval model deployed in SfM pipeline for accelerating image matching, the full augmentation then requires only $\scriptsize{\sim}5\%$ time cost compared with the raw local feature description. Virtually, the proposed framework allows flexible integration and reuse of other visual components to achieve system-level efficiency, such as saliency or segmentation masks, and thus has large rooms for future improvements.

\begin{table}[h]
	\caption{The computational cost of proposed framework, evaluated on 10k keypoints from an $896\times 896$ image. The inference time is estimated on an NVIDIA GTX 1080 GPU.}
	\label{tab:cost}
	\resizebox{0.48\textwidth}{!}{
	\begin{tabular}{ccc|ccc}
		\Xhline{1pt}
		 &\multicolumn{2}{c}{\textit{Preparation}}  & \multicolumn{3}{c}{\textit{Augmentation}}                                               \\ \hline
		 &\textbf{local feat.} & \textbf{regional feat.} & \textbf{geo. context} & \textbf{vis. context} & \textbf{multi-context} \\
		 Time (ms) &351               & 49               & 5 & 14 & 18     \\
		 FLOPs (B) &802.9               & 123.4               & 1.7 & 13.9 & 15.7     \\
		 Params (M) & 2.4               & 24.5               & $<$0.1 & 3.1 & 3.2     \\ \Xhline{1pt}
	\end{tabular}
}
\end{table}

\smallskip\noindent\textbf{End-to-end training.}
For ablation purposes, the parameters of base local and regional models are previously fixed in the training, and we here provide further studies about the efficacy of an end-to-end training scheme.

In the first setting, we freeze only the regional model and train
from scratch with Eq.~\ref{equ:tot_loss} on the
augmented feature. As a result, the performance is notably improved from \textbf{67.14 to 67.53}, and \textbf{76.42 to 77.20} for \emph{i/v} sequences of HPSequences, compared with \emph{ContextDesc+} in Tab.~\ref{tab:ablation}.

In the second setting, we further end-to-end train with the regional model, which is additionally optimized by a standard cross-entropy classification loss as in~\cite{noh2017large} for simplicity (see Appendix~{A.\color{red}1} for details). Although several loss balancing strategies have been experimented, we did not observe a consistent improvement for final matching performance, which we ascribe to the substantial challenge posing by multi-task learning. Thus, we currently recommend a separate training for the regional model, and look forward to
an improved solution in the future.

\section{Conclusion}

In contrast to current trends, we have addressed the importance of introducing \emph{context awareness} to augment local feature descriptors. The proposed framework takes keypoint location, raw local and high-level regional feature as input, from which two types of context are encoded: \emph{geometric} and \emph{visual} context, while the training adopts a novel N-pair loss that is self-adaptive and parameter-tuning free. We have conducted extensive evaluations on diversified and large-scale datasets, and demonstrate remarkable improvements over the state of the art, meanwhile showing the strong generalization and practicability in real applications.

\paragraph{Acknowledgment.} This work is supported by Hong Kong RGC GRF 16203518, T22-603/15N, ITC PSKL12EG02. We thank the support of Google Cloud Platform.

\clearpage

{\small
	\bibliographystyle{ieee}
	\bibliography{egbib}
}

\clearpage
\section*{A. Supplementary appendix}
\subsection*{A.1 Implementation details}

\paragraph{Network details.}
In terms of the matchability predictor, we construct 4-layer MLPs whose output node numbers are 128, 32, 32, 1, respectively. The visual context encoder is composed of two 2-layer MLPs, located before/after the concatenation with raw local features. We insert context normalization only into the former MLPs, while  insertion in the latter one is observed to harm the performance.

\paragraph{Performance of the retrieval model.}
The retrieval model is trained on \emph{Google-Landmarks Dataset}~\cite{noh2017large}, which contains more than 1M landmark images. In our experiments, we have compared different networks for the retrieval performance. In brief, ResNet-101 is slightly better than ResNet-50, while VGG and AlexNet are notably underperforming. We choose ResNet-50 for better tradeoffs in memory and speed.

Instead of adopting the training scheme in~\cite{radenovic2016cnn}, we find that the model pretrained on landmark classification task (containing 15K classes) as in~\cite{noh2017large} suffices to produce satisfactory results in practice, and avoids difficulties on preparing training data for Siamese networks or hard negative mining with complex heuristics. We have evaluated the retrieval model with MAC aggregation on standard Oxford buildings dataset~\cite{philbin2007object}, where we obtain mAP of 0.83, on par with~\cite{radenovic2016cnn} of 0.80.

\paragraph{Keypoint coordinate augmentation.}
Similar to~\cite{detone2016deep}, we choose to use 4-point parameterization to represent the homography as follows:
\begin{equation*}
H_{4point}= 
\left\{
\begin{matrix}
u_1 + \triangle u_1 & v_1 + \triangle v_1 \\
u_2 + \triangle u_2 & v_2 + \triangle v_2 \\
u_3 + \triangle u_3 & v_3 + \triangle v_3 \\
u_4 + \triangle u_4 & v_4 + \triangle v_4
\end{matrix}
\right\},
\end{equation*}
where $(u_1, v_1), (u_2, v_2), (u_3, v_3), (u_4, v_4)$ are four corner points at $(-1, 1), (1, 1), (-1, -1), (1, -1)$, and $\triangle u_i, \triangle v_i$ are random variables between $(-0.5, 0.5)$. One can easily convert $H_{4point}$ to a standard $3\times3$ homography by, e.g., normalized
Direct Linear Transform (DLT) algorithm. In our implementation, we apply the random homography on each keypoint coordinate set before feeding it into the geometric context encoder.

\begin{figure}[th]
	\centering 
	\includegraphics[width=0.25\textwidth]{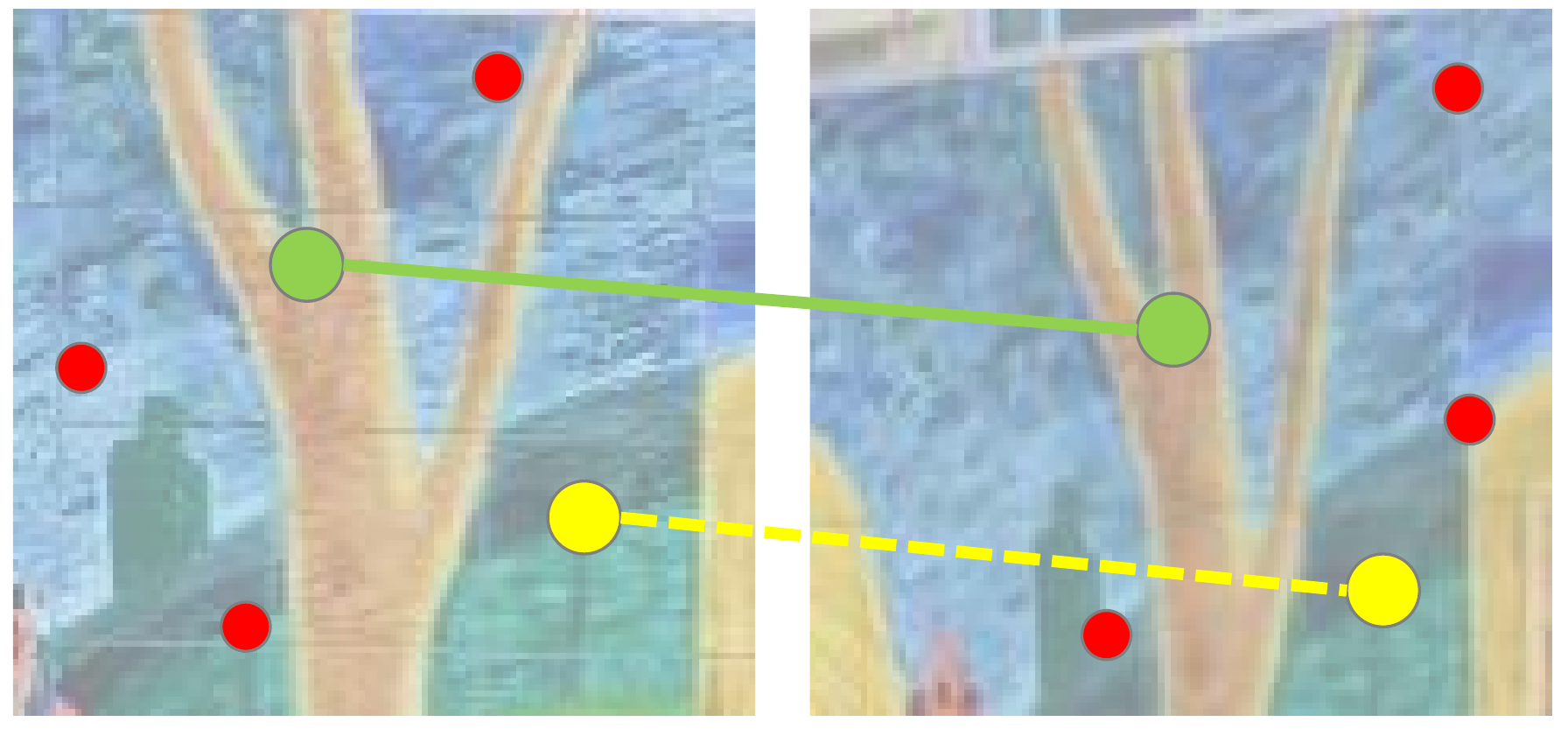}
	\vspace{-2mm}
	\caption{We divide keypoints after SfM into three categories: matchable ({\color{ForestGreen}green}), undiscovered ({\color{Goldenrod}yellow}) and unrepeatable ({\color{red}red}), and aim to perform a complete simulation in training that incorporates all three types of keypoints.}
	\label{fig:kpt_type}
\end{figure}

\paragraph{Learning with noisy keypoints.}
\label{sec:learning_scheme}
The training of proposed framework, apparently, needs to be conducted between image pairs instead of isolated patches, since we also take keypoint coordinates as input. Such data organization is referred to as simulating image matching in~\cite{Luo_2018_ECCV}. However, the simulation in~\cite{Luo_2018_ECCV} is not complete, as it considers only keypoints that have successfully established correspondences, whereas in real situation, only a subset of keypoints is repeatable in other images. In practice, as illustrated in Fig.~\ref{fig:kpt_type}, we divide keypoints obtained from SfM as in ~\cite{yi2016lift,Luo_2018_ECCV} into three categories: i) \emph{Matchable}: repeatable and verified by SfM; ii) \emph{Undiscovered}: repeatable but did not survive the SfM. iii) \emph{Unrepeatable}: unable to be re-detected in other images. 

In the training, we randomly sample a number of matchable keypoints as well as some undiscovered and unrepeatable keypoints (denoted as \emph{noisy keypoints}), in order to have a complete simulation that is necessary to acquire strong generalization ability. Otherwise, the training will consider only ideal setting with all matchable keypoints, which is inconsistent with real applications.

To incorporate with the above training strategy, we need to make some adaptation  on the loss formulation. Formally, given index sets $C_m=\{i_1, . . . , i_{K_m}\}$ and $C_n=\{i_1, . . . , i_{K_n}\}$, where $K_m$ and $K_n$ are numbers of matchable and noisy keypoints for an image pair, the losses of Eq.~\ref{eq:quad_loss} and Eq.~\ref{eq:npair} are now rewritten as:
\begin{equation}
\label{eq:new_quad_loss}
\begin{aligned}
& \mathcal{L'}_{quad} = \\
& \frac{1}{K_m(K_m-1)}\sum_{i, j\in C_m, i\neq j}\max(0, 1 - R(\bm{f}_1^i, \bm{f}_1^j, \bm{f}_2^i, \bm{f}_2^j)),
\end{aligned}
\end{equation}
and
\begin{equation}
\mathcal{L'}_{N\text{-}pair}=-\frac{1}{2}(\sum_{i\in{C_m}}\log{s_{ii}^r} + \sum_{i\in{C_m}}\log{s_{ii}^c}).
\end{equation}
Subsequently, adding noisy keypoints will first influence the encoding of geometric context, posing a harder training settings and playing a key role in order to acquire the invariance properties. Second, it will influence the computation of $\mathcal{L}_{N\text{-}pair}$, as those keypoints will be all cross-paired as negative samples. It also enables us to increase the pair number in each batch, i.e., 1024 in our implementation compared with 64 in GeoDesc~\cite{Luo_2018_ECCV}, which boost the effectiveness of N-pair loss as observed in~\cite{mishchuk2017working}. 

\paragraph{Further joint processing in aggregation step.}
In this work, as in Sec.~\ref{sec:agg}, we simply sum and normalize the cross-modality features for aggregation. Meanwhile, we have also attempted to make this module learnable by concatenating and feeding the features to several fully-connected layers. However, the experimental results showed a considerable performance decrease from such choice, i.e., 2 points decrease on HPatches even compared with the base model, GeoDesc. Our observation is that the raw local features are supposed to be preserved as much as possible, and a learnable
aggregation would result in over-parameterization and inability to represent the local detail. 

\subsection*{A.2 Training with softmax temperature}
\begin{figure}[h]
	\centering 	
	\includegraphics[width=0.48\textwidth]{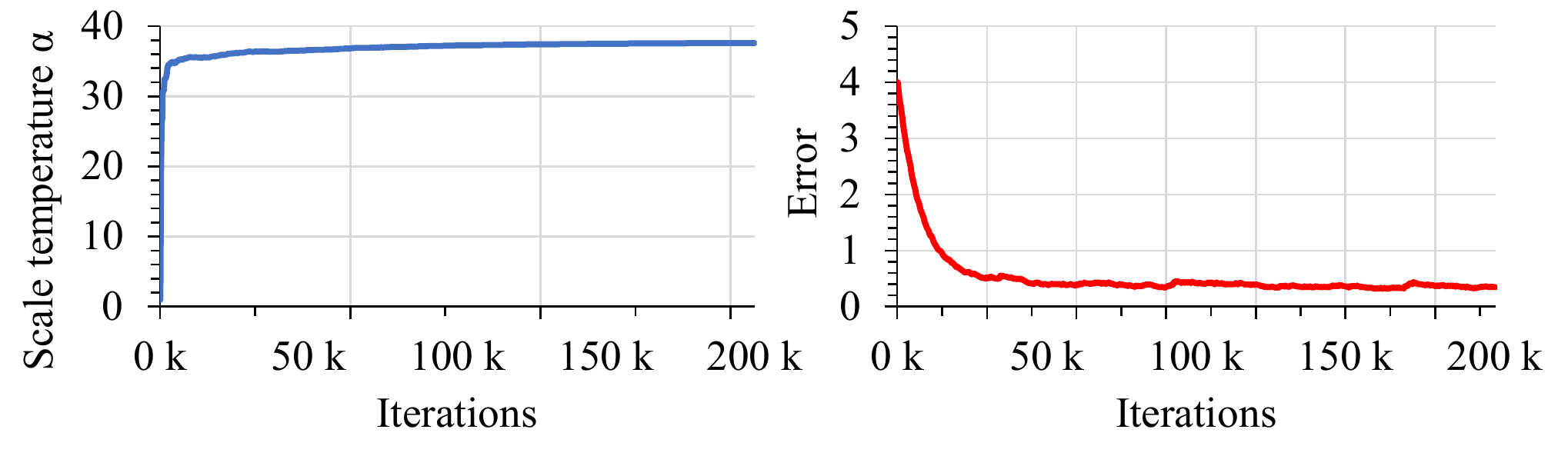}
	\vspace{-4mm}
	\caption{Left: the growth of scale temperature. Right: the respective decrease of loss.}
	\label{fig:softmax_temperature}
\end{figure}

We plot the growth of softmax temperature and its respective loss decrease in Fig.~\ref{fig:softmax_temperature}. As can be seen, the softmax temperature fast grows at the beginning and gradually converges to a constant value, $\scriptsize{\sim}38$. As mentioned in Sec.~\ref{sec:loss}, the softmax temperature is regularized with the same network weight decay, whereas we have observed that eschewing the regularization does not harm the performance, but results in a larger temperature value, i.e., $\scriptsize{\sim}42$. 

\subsection*{A.3 Ratio test}
In previous experiments on image matching, we did not apply any outlier rejection (e.g., mutual check, ratio test~\cite{lowe2004distinctive}) for all methods for fair comparison, whereas the early outlier rejection is critical and necessary to later geometry computation, e.g., recovering camera pose. In particular, ratio test~\cite{lowe2004distinctive} has demonstrated remarkable success, we thus follow the practice in~\cite{Luo_2018_ECCV} to determine the ratio criteria of the proposed augmented descriptor. Specifically, given \emph{\# Correct Matches} defined in Sec.~\ref{sec:metric}, we test on HPSequence~\cite{balntas2017hpatches} and aim to find a proper ratio that achieves \emph{Precision} = \emph{\# Putative Matches} / \emph{\# Correct Matches} similar to SIFT. As a result, we choose $0.89$ for the proposed descriptor.

\begin{table}[h]
	\centering
	\caption{Pose evaluation on wild datasets with ratio test applied: \emph{indoor} SUN3D~\cite{xiao2013sun3d} and \emph{outdoor} YFCC100M~\cite{thomee2015yfcc100m}.}
	\begin{tabular}{cccc}
		\Xhline{1pt}
		& \textbf{SIFT\cite{lowe2004distinctive}} & \textbf{GeoDesc~\cite{Luo_2018_ECCV}} & \textbf{Ours} \\ \hline
		\textit{ratio criteria} & 0.80 & 0.89 & 0.89 \\ \hline
		\multicolumn{4}{c}{\textit{mAP of pose (error threshold 20\textdegree)}}                                   \\ \hline
		\textit{indoor}      &  37.4                &     41.8         & \textbf{42.9}                    \\
		\textit{outdoor} &    17.9              &          20.5      &      \textbf{22.5}               \\ \Xhline{1pt}
	\end{tabular}
	\label{tab:wild_pose}
	
\end{table}

To demonstrate the efficacy of the obtained ratio, we evaluate on the wild indoor/outdoor data~\cite{xiao2013sun3d,thomee2015yfcc100m} with an error metric of relative camera pose accuracy. Following the protocols defined in~\cite{yi2018learning}, we use mean average precision (mAP) of a certain threshold (e.g., 20\textdegree) to quantify the error of rotation and translation. For comparison, we use ratio criteria of 0.80 for SIFT~\cite{lowe2004distinctive} and 0.89 for GeoDesc~\cite{Luo_2018_ECCV}, and present evaluation results in Tab.~\ref{tab:wild_pose}, which demonstrates consistent improvements with proper outlier rejection.

\subsection*{A.4 Different domain sizes}
Somewhat counter-intuitively, the CS structure improves marginally on image matching tasks as reported in Tab.~\ref{tab:ablation}. To further study this phenomenon, we compare the patch sampling from different domain sizes, including the original SIFT's ($1\times$) as used in previous experiments, half  ($0.5\times$) or double ($2\times$) sizes. We also compare the aggregation of multiple sizes, i.e., the original and halved $(1+0.5)\times$ or the original and doubled $(1+2)\times$. Instead of concatenating features as used by CS structure, we apply simple summing-and-normalizing aggregation in Sec.~\ref{sec:agg} to avoid increasing the dimensionality. 

We experiments with our \emph{ContextDesc+} model in Tab.~\ref{tab:ablation}, and present the comparison results with different domain sizes in Tab.~\ref{tab:domain}. As can be seen, when only single size is adopted, the original `\emph{$1\times$}' performs best as being consistent with the training. In addition, when combining a larger size ($(1+2)\times$), we can further boost the proposed method by a considerable margin, yet leading to excessive computational cost and doubling the inference time. In practice, the aggregation with different domain sizes is compatible with the proposed framework, and can be applicable when high accuracy is in demand.

\begin{table}[h]
	\centering
	\caption{The efficacy of extracting local features from different domain sizes.}
	\begin{tabular}{l|c|c}
		\Xhline{1pt}
		\multicolumn{1}{c}{\textbf{domain size}} & \multicolumn{2}{c}{\textbf{Recall \emph{i/v}}} \\ \hline
		\textit{$0.5\times$}                            &  61.59                &     69.79             \\
		\textit{$2\times$}                              &    62.84         &      71.86       \\ 
		\textit{$1\times$} (\textbf{ContextDesc+})                             & \textbf{67.14}            & \textbf{76.42}            \\ \hline
		\textit{$(1+0.5)\times$}                       &     67.31             &     76.16             \\
		\textit{$(1+2)\times$}                         &  \textbf{67.74} & \textbf{77.51}                           \\ \Xhline{1pt}
	\end{tabular}
	\label{tab:domain}
\end{table}

\begin{figure*}[th]
	\centering 	
	\includegraphics[width=\textwidth]{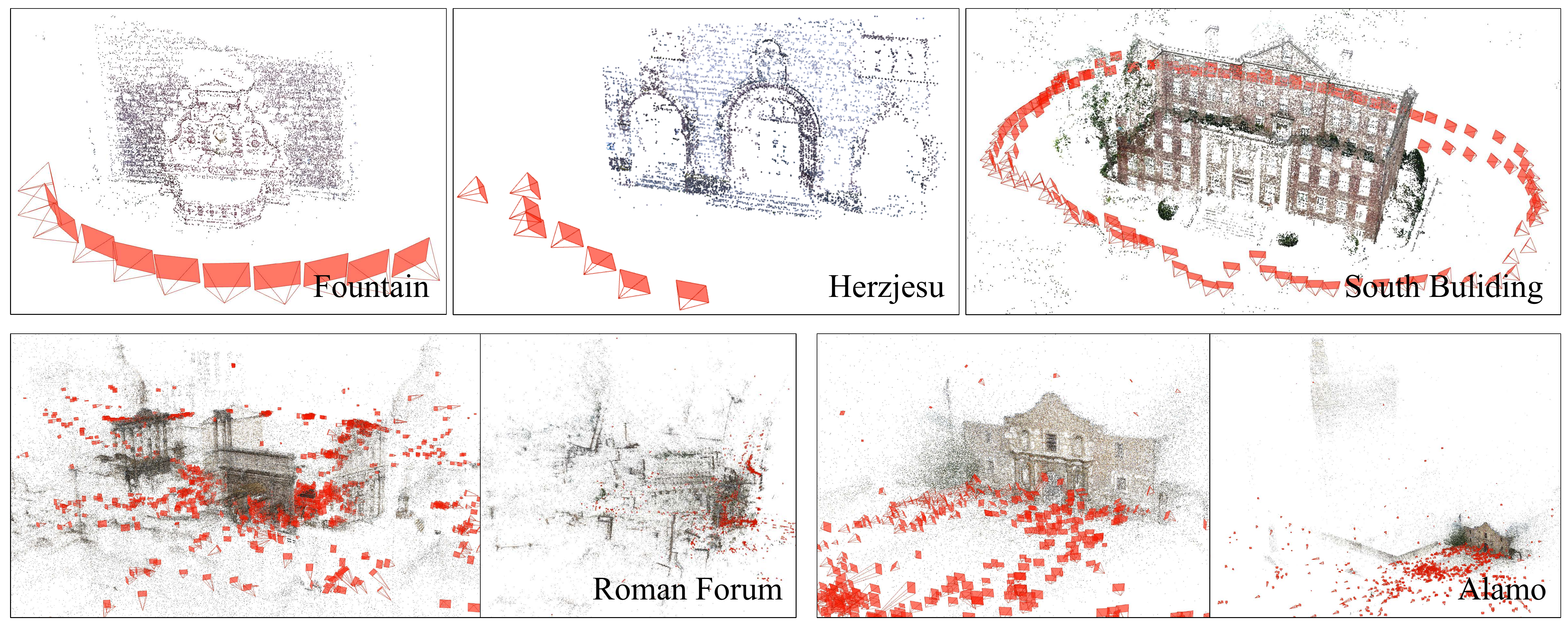}
	\vspace{-4mm}
	\caption{Visualizations of SfM results of Sec.~\ref{sec:generalization} from the proposed augmented feature.}
	\label{fig:sfm_vis}
\end{figure*}

\subsection*{A.5 Invariance to density change}
\begin{figure}[th]
	\centering 
	\includegraphics[width=0.48\textwidth]{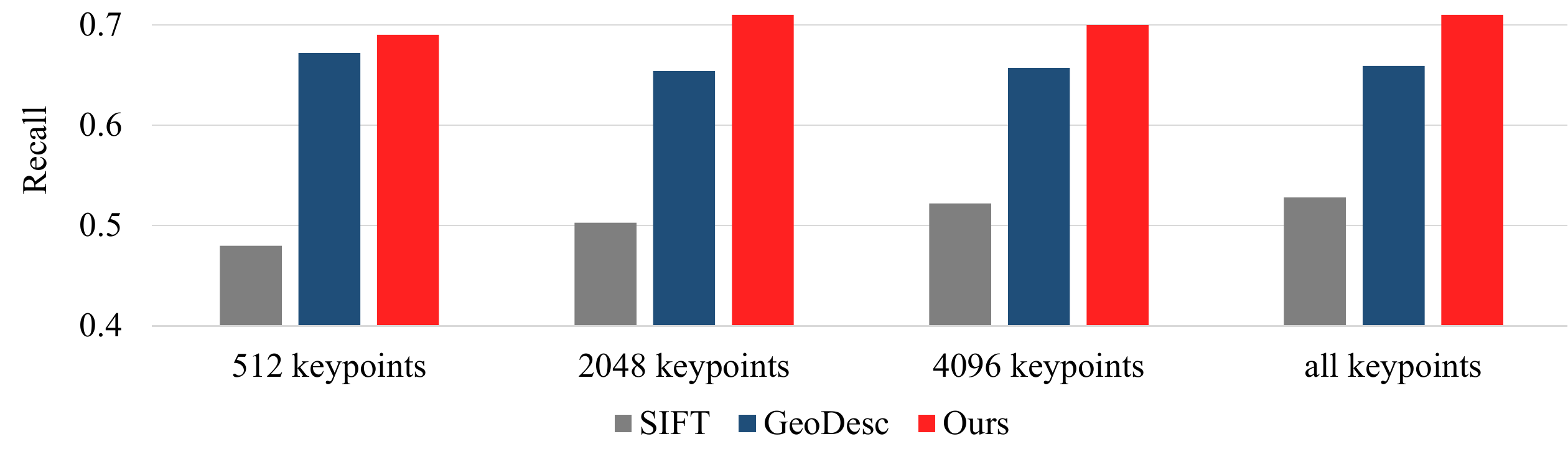}
	\caption{The performance of proposed augmentation scheme regarding density change of keypoints.}
	\label{fig:density}
\end{figure}

We further demonstrate the robustness regarding density change on HPSequences~\cite{balntas2017hpatches}, of which images are feature-rich and have keypoints up to 15k. Beside of sampling keypoints of different numbers, we consider a more challenging case where \emph{all detected keypoints} are used. As presented in Fig.~\ref{fig:density}, the proposed method delivers consistent improvements in terms of all cases, which demonstrates the reliable invariance property acquired by context encoders.

\subsection*{A.6. Efficacy of the matchability predictor}
To better interpret the functionality of the proposed matchability predictor in Sec.~\ref{sec:gc}, we quantitatively evaluate its performance being used as a keypoint detector. Following~\cite{savinov2017quad}, we apply the matchability predictor onto the entire image, then select $2048$ top responses after NMS as keypoints.  whose performance is measured by \emph{Repeatability}. Compared with SIFT detector, the results are improved from \textbf{32.81 to 37.93} and \textbf{25.53 to 26.34} on \emph{i/v} sequences of HPatches. While the detector performance is not the focus of this paper, we believe that by adopting more advanced techniques, this module will potentially benefit to the \emph{joint training} of keypoint detector and descriptor, and have large rooms for future improvements.

\subsection*{A.7 Application on image retrieval}
We use an open-source implementation of VocabTree\footnote{https://github.com/hlzz/libvot}~\cite{shen2016graph} for evaluation image retrieval performance, and compare SIFT~\cite{lowe2004distinctive}, GeoDesc~\cite{Luo_2018_ECCV} and the proposed ContextDesc. The mAPs on Paris dataset~\cite{philbin2008lost} from three competitors are \textbf{49.89, 53.84 and 61.29}, while on Oxford buildings~\cite{philbin2007object} are \textbf{47.27, 53.29 and 61.64}. By re-ranking the top-100 by spatial verification~\cite{philbin2007object}, the mAPs on Paris are improved to \textbf{52.23, 55.02 and 64.53}, while on Oxford are \textbf{51.64 , 54.98 and 65.03}. The experimental results effectively demonstrate the superiority of the proposed method.

\subsection*{A.8 More visualizations}
\begin{figure*}[th]
	\centering 	
	\includegraphics[width=0.97\textwidth]{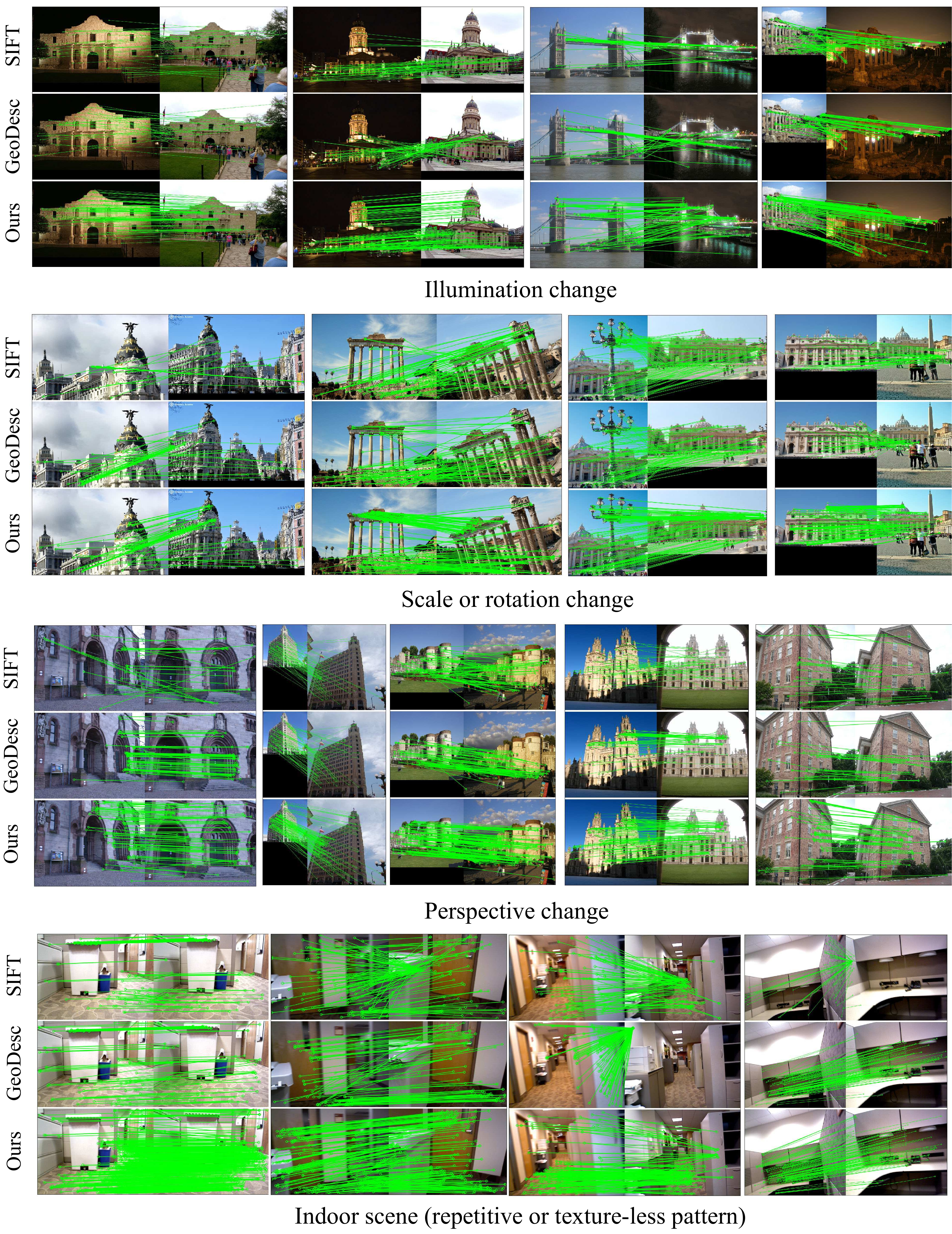}
	\caption{Image matching results after RANSAC. From top to bottom: SIFT, GeoDesc and proposed augmented feature.}
	\label{fig:supp_vis}
\end{figure*}

We have provided more visualizations of previous experiments in Fig.~\ref{fig:sfm_vis} (SfM results in Sec.~\ref{sec:generalization}) and Fig.~\ref{fig:supp_vis} (image matching results w.r.t different image transformations).

\end{document}